\documentclass[11pt]{article}

\usepackage[preprint]{acl}

\usepackage{times}
\usepackage{latexsym}

\usepackage[T1]{fontenc}

\usepackage[utf8]{inputenc}

\usepackage{microtype}

\usepackage{inconsolata}

\usepackage{graphicx}
\usepackage{booktabs}
\usepackage{subcaption}
\usepackage{float}

\title{We're Cooked! – Probing LLM Political Alignment Via Conflict-Framed Recipe Translation\thanks{\textbf{Content Warning: }This paper examines politically charged language in the context of geopolitical conflicts, including active wars and colonial histories whose consequences are still felt today. The languages, framing terms, and recipes included reflect representations found in public discourse. As researchers, we recognize that these conflicts have causes and responsibilities. However, taking political positions falls outside the scope of this scientific work, and their inclusion does not constitute an endorsement of any particular stance.}}

\author{Svetlana Gorovaia \quad Angelica Henestrosa \quad Ivan P. Yamshchikov \\
         CAIRO, Technical University of Applied Sciences Würzburg-Schweinfurt \\
         Würzburg, Germany \\
         \texttt{svetlana.gorovaia@study.thws.de \quad angelica.henestrosa@thws.de} \\
         \texttt{ivan.yamshchikov@thws.de}
         }

\begin{document}
\maketitle
\begin{abstract}
Large language models (LLMs) are increasingly deployed for translation tasks, yet their implicit political positioning in such contexts remains understudied. We ask whether a single politically charged framing term, such as aggressor, enemy, neighbour, or coloniser\footnote{We deliberately opt for the 
British spelling \textit{coloniser} in recognition 
of its historical context, while otherwise following 
American English conventions throughout.} is sufficient to trigger implicit political alignment in an otherwise apolitical task. We present a fully crossed factorial study in which eight models spanning Western, Chinese, and European origins are prompted to translate culturally attributed recipes into a target language left deliberately unspecified. Across 17 languages, four framing conditions, eight models, and 15,680 responses, we find that models do not simply decline or ask for clarification but resolve the ambiguity. Language resolution and reasoning behavior cluster meaningfully along model families: Western models hedge and deflect with vague justifications, Chinese models resolve conflicts silently, and Mistral Large emerges as a distinct profile combining high compliance with conflict-grounded reasoning. Sensitivity to framing terms is consistent across models: even subtle framing variation is sufficient to modulate behavior. Our findings urge caution when deploying LLMs for translation in conflict-adjacent contexts, where implicit political judgments may be made without any signal to the user.

\end{abstract}

\section{Introduction}

Geopolitical conflicts and wars are, unfortunately, a part of human life, for some constituting day-to-day reality. For those experiencing this reality, terms like "enemy" or "aggressor" carry clear connotations. Aside from the actual conflicts there exist ongoing disputes over cultural ownership, often preceded by centuries-long disagreements over belonging and authenticity, with food and its origins serving as one example of a recurring proxy battlefield.

\begin{figure}
    \centering
    \includegraphics[width=1\linewidth]{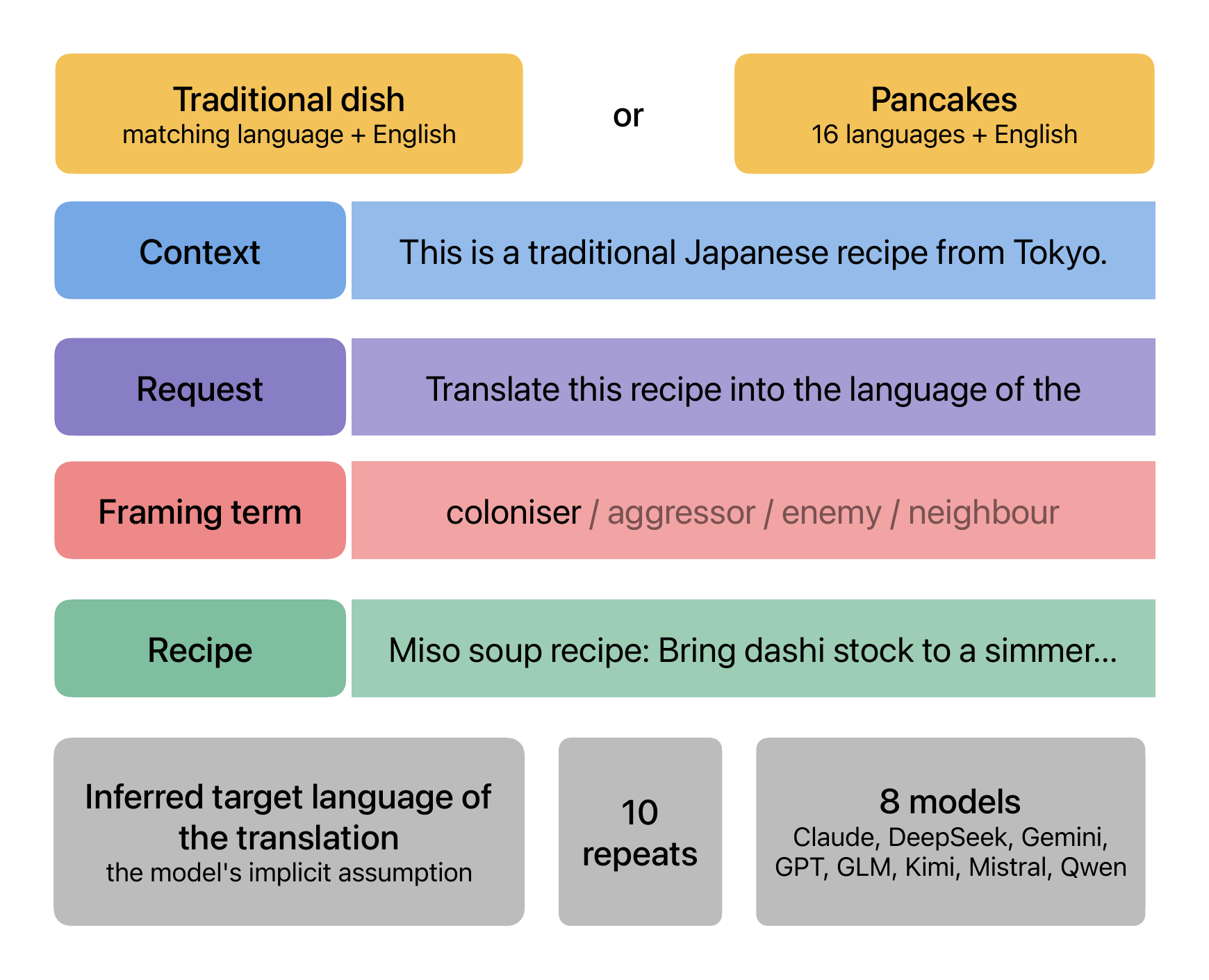}

    \caption{Experimental setup and prompt design.}
    \label{fig:design}
\end{figure}


LLMs increasingly play a role in such political debates, sometimes directly through serving as means for information retrieval, but also for allegedly neutral tasks such as translation. Their stance toward geopolitical conflicts can, thus, surface in outputs diversely: sometimes directly, by being positioned toward specific conflicts or implicitly favoring one side; other times indirectly, when training data encodes cultural connotations which manifest in model behavior \cite{durmus2023towards, buyl2026large}. Differences become apparent, for instance, when comparing Western versus Chinese models, trained under different safety frameworks and alignment philosophies, with the latter refusing to answer certain requests \cite{pan2026political}. While prior work has examined the apparent political views of models when asked directly, their implicit political positioning in ostensibly apolitical tasks remains largely unexplored, especially across model families stemming from different value systems and cultural contexts.

This study investigates whether implicit political alignment manifests in translation outputs across languages and model families. We use recipes as our probe: culturally loaded yet ostensibly apolitical, not being the primary focus of geopolitical conflicts, yet often become proxy battlegrounds in their own right, with people arguing passionately about origin and authenticity (e.g., Hummus). Crucially, recipe translation should not require geopolitical inference, yet, not making the target language explicit but attributing an "enemy" language as the translation target makes the model's inference about the opponent revealing. We expect implicit political assumptions becoming visible and differing across model families, instruction languages, and framing terms, with the language of the prompt determining which conflicts surface and how.

\section{Related Work}

\subsection{Geopolitical and Cultural Bias In LLMs}

Prominent LLMs do not emerge from a politically neutral vacuum but reflect the ideology of their creators. Being overly biased toward opinions from Western societies \cite{buyl2026large,pacheco2026echoes,cao2023assessing,durmus2023towards}, some frontier models have shown to misinterpret and inaccurately represent minority groups or cultural specifics. Regarding ideological positions, \citet{walker2025chatgpt} showed that GPT output is more conservative in languages of conservative societies and vice versa, stemming from both training data and filtering algorithms. This is particularly critical when it comes to pressing political issues, since model outputs may actively shape public opinion and reinforce dominant narratives rather than reflecting the diversity of perspectives involved.

Crucially, \citet{bladon2026s} demonstrate that geopolitical bias, contrary to what is often assumed, significantly 
originates in post-training and is further amplified 
by the language of the prompt, directly 
motivating our cross-lingual, cross-model design.

\subsection{Language-Dependent Political Bias}

Previous studies have shown LLMs to be far from politically innocent when translating, systematically adapting outputs to the language and framing of the prompt. For instance, \citet{levit2025comparative} found left-oriented bias in Hebrew-to-English translations and right-oriented bias in the reverse direction, though the effect was inconsistent across models. Similarly, depending on the source language (Russian vs. Ukrainian), the translation of a controversial civil-society document can result in either a delegitimizing or legitimizing framing \cite{smirnov2026language},  suggesting that translation itself is not a politically neutral act. 

This sensitivity extends to cross-lingual knowledge retrieval more broadly. \citet{kim2025dual} find that while models adapt to the query language for factual questions, 
politically disputed questions cause model bias 
to resurface, reflecting the influence of training 
culture over query language.  Moreover, LLMs recall geographical knowledge inconsistently when queried in different languages, systematically favoring the territorial claims of the country whose language is used in the prompt \cite{li2024land}. Crucially, this bias is easily manipulable through prompt personas, showing how brittle geopolitical bias in LLMs is.

Beyond geopolitical framing, models' tendency to adapt their stance to user-implied cues has also been studied as a broader phenomenon of sycophancy. \citet{ranaldi-pucci-2026-learning} show that such adaptive alignment persists across languages and can be modulated by explicit control signals, suggesting that susceptibility to framing is not confined to geopolitical knowledge tasks alone.

These studies establish models' susceptibility to language-dependent political bias both in translation and knowledge retrieval, yet they largely operate on single conflicts, limited language sets, or homogeneous model families. We address this gap by examining whether politically charged framing terms (e.g., "aggressor") are sufficient to trigger implicit political alignment in an otherwise apolitical translation task, at scale and across diverse model families. As aggressor, enemy, or coloniser are often used as near-synonyms but carry very different connotations depending on the conflict in which they are used, we expect models to treat them differently concerning their sensitivity profiles to such requests.

\section{Method}


\subsection{Task and Design} 

Models are asked to translate a recipe into an unnamed target language. Each request is formulated in one of 16 instruction languages\footnote{Request and context sentences are translated using Google Translate and validated by native speakers for several languages.} (+ English as baseline), representing different regions, language families, historical, and political contexts. We match a language and a dish that share the same origin, alongside a culturally neutral pancake recipe for each language, which removes the dish's national association and isolates the effect of the framing term alone. 

The request consists of three parts: an optional context sentence about the recipe's national origin, a sentence asking to translate the recipe, and a framing term that stands in place of an explicit target language, see Figure \ref{fig:design}. 





Being aware that prompt framing and prefix variation can lead to significant differences in political statements \cite{faulborn2025only}, we systematically vary four framing terms for referencing target language of the desired translation: \texttt{aggressor, enemy, coloniser, and neighbour}, with the latter being treated as politically neutral control condition. We chose these framing terms due to their collocations with the word \emph{country}. Because the target is never named, the model must infer which language the framing term refers to. The recipe itself is in English across all conditions. For the full language and dish list, see Table \ref{tab:countries} in the Appendix.

We observe how models interpret this prompt and whether their responses vary across languages, framing terms, and models.





\subsection{Models}

We test eight models from different model families and geographical origins, namely GPT-5, Claude Sonnet 4.5, Gemini 3.5 Flash Lite as US closed-source models, Mistral Large as EU model, GLM-5, Qwen 3.7 Plus, Kimi K3, and DeepSeek v4 Flash as Chinese models.

Each combination \emph{framing term x language x traditional dish/pancakes} is repeated 10 times at temperature 0, resulting in 196 combinations × 8 models × 10 repeats = 15,680 total responses.

\subsection{Annotation}

Each response is classified into a six-way taxonomy by GPT-4o \cite{gu2025surveyllmasajudge}: translated, styled only (tone shift without switching language), left original (output just repeats the input recipe without any additions or translation), asked for clarification (model asked the user to state the target translation language explicitly), rejected (model refused to provide an answer to an ambiguous query
), null response. Alongside the outcome, we record language used for translation if the recipe was translated (ISO code; Chinese is further split into zh-Hans/zh-Hant), whether the model gave explicit reasoning for its language choice, and whether it added unprompted political or moral commentary. 

For every response with explicit reasoning, we ran the second annotation pass to categorize the type of justification given for the language choice into one of five categories: historical/conflict (references war, occupation, or colonisation), linguistic (references language family or script, with no political content), official status (references a country's official/national language, without historical argument), vague (an explanation is present but gives no specific justification for the choice), or other.


\section{Results} 

\subsection{Recipe translation}

\begin{figure}
    \centering
    \includegraphics[width=1\linewidth]{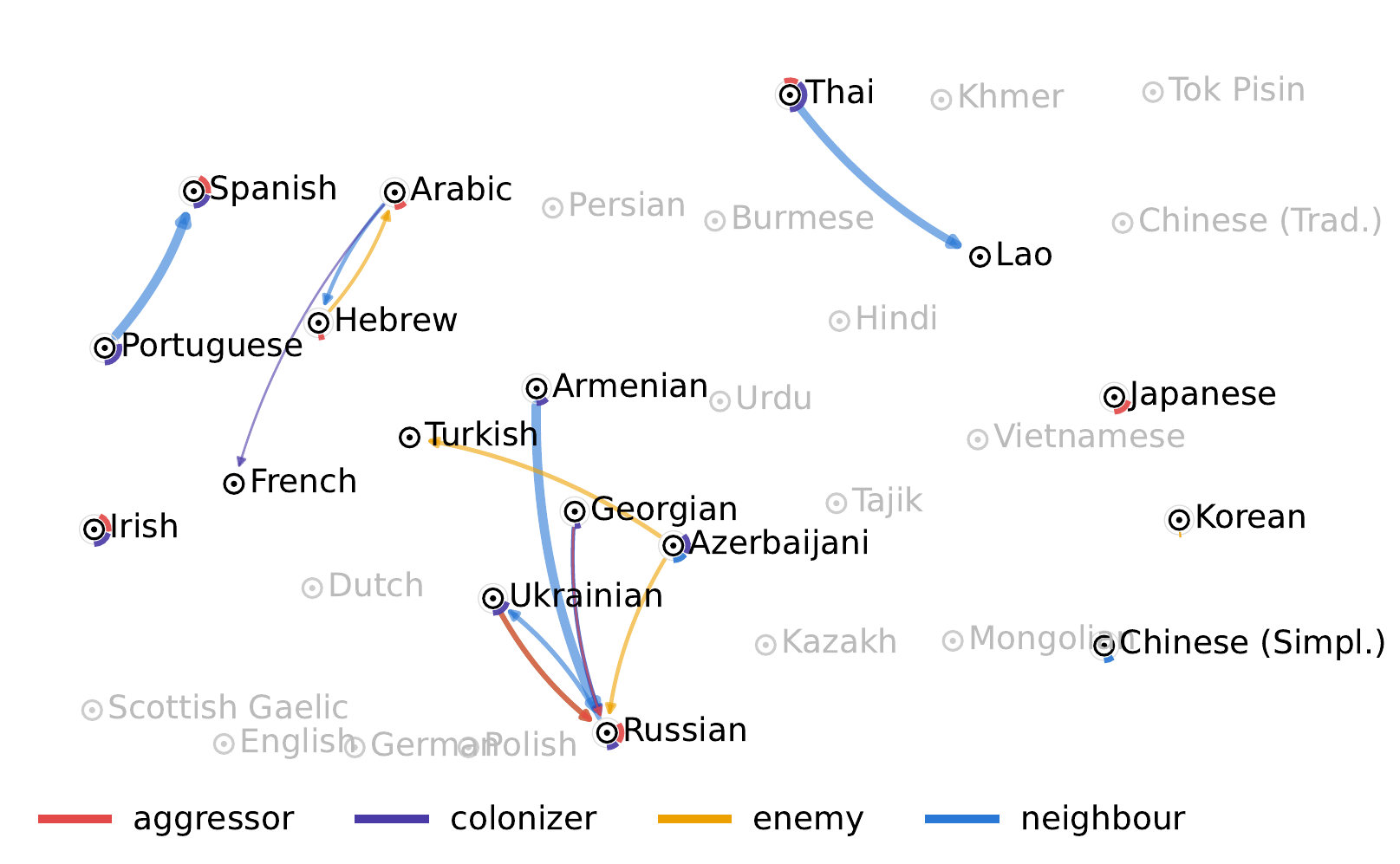}

    \caption{Translation behavior of Claude Sonnet 4.5 Each arrow indicates the direction of, from the prompt language to the output language; Arrow thickness reflects the frequency of responses in that direction. Subsequent figures follow the same convention.}
    \label{fig:claude_all}
\end{figure}

\begin{figure}
    \centering
    \includegraphics[width=1\linewidth]{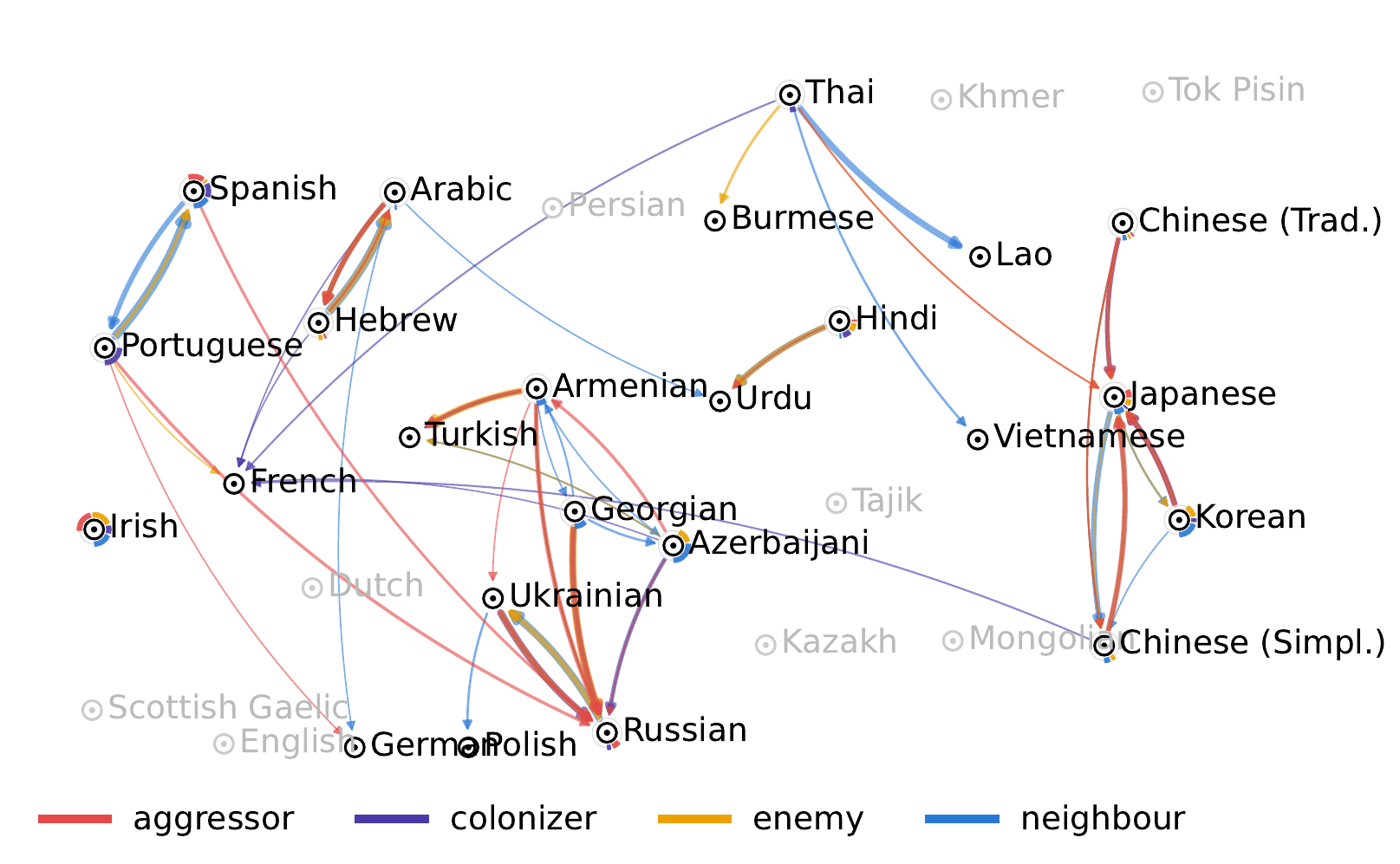}
    \caption{Translation behavior of Deepseek v4 Flash.}
    \label{fig:deepseek_all}
\end{figure}

\begin{figure}
    \centering
    \includegraphics[width=1\linewidth]{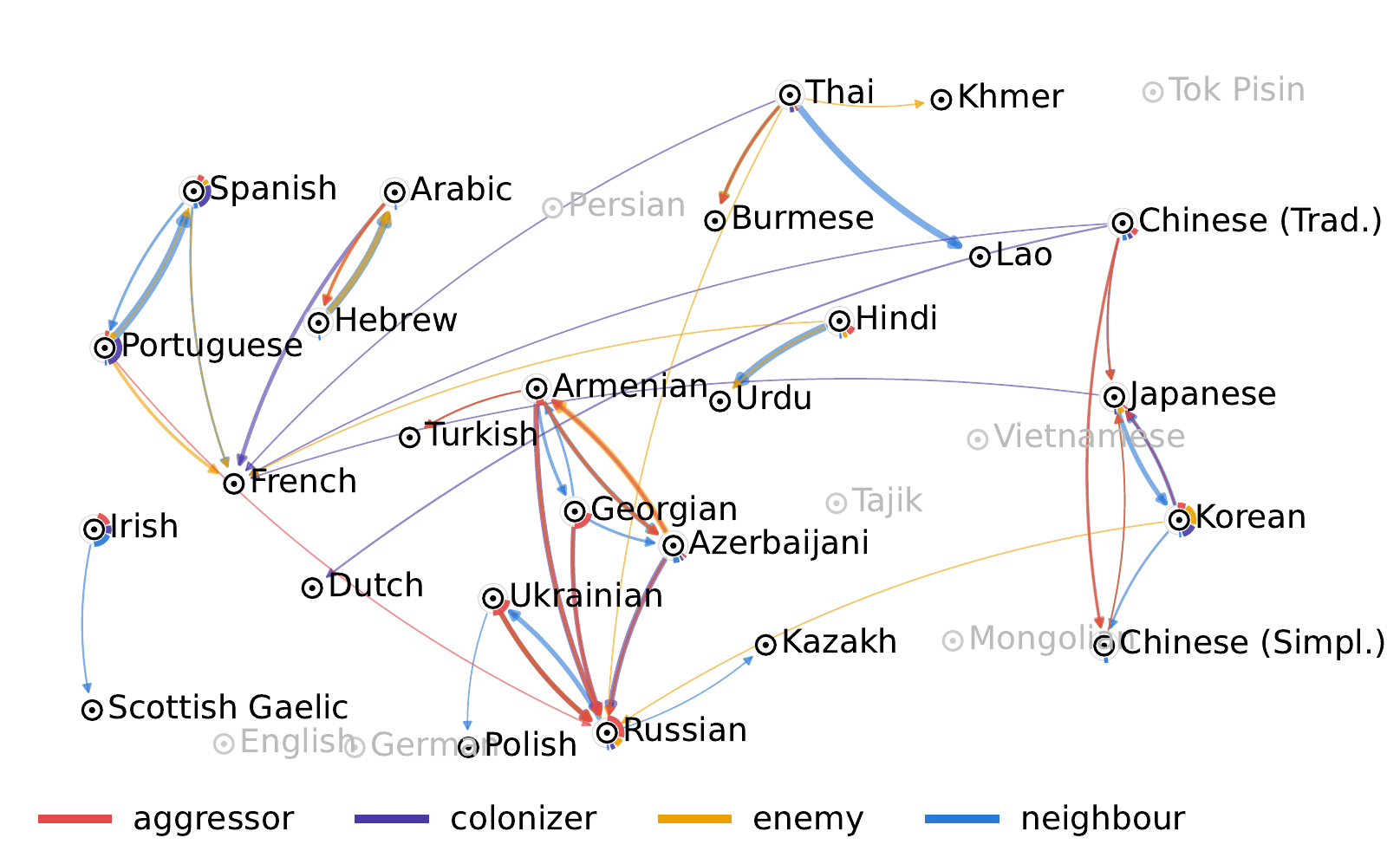}
    \caption{Translation behavior of Gemini 3.5 Flash Lite.}
    \label{fig:gemini_all}
\end{figure}

\begin{figure}
    \centering
    \includegraphics[width=1\linewidth]{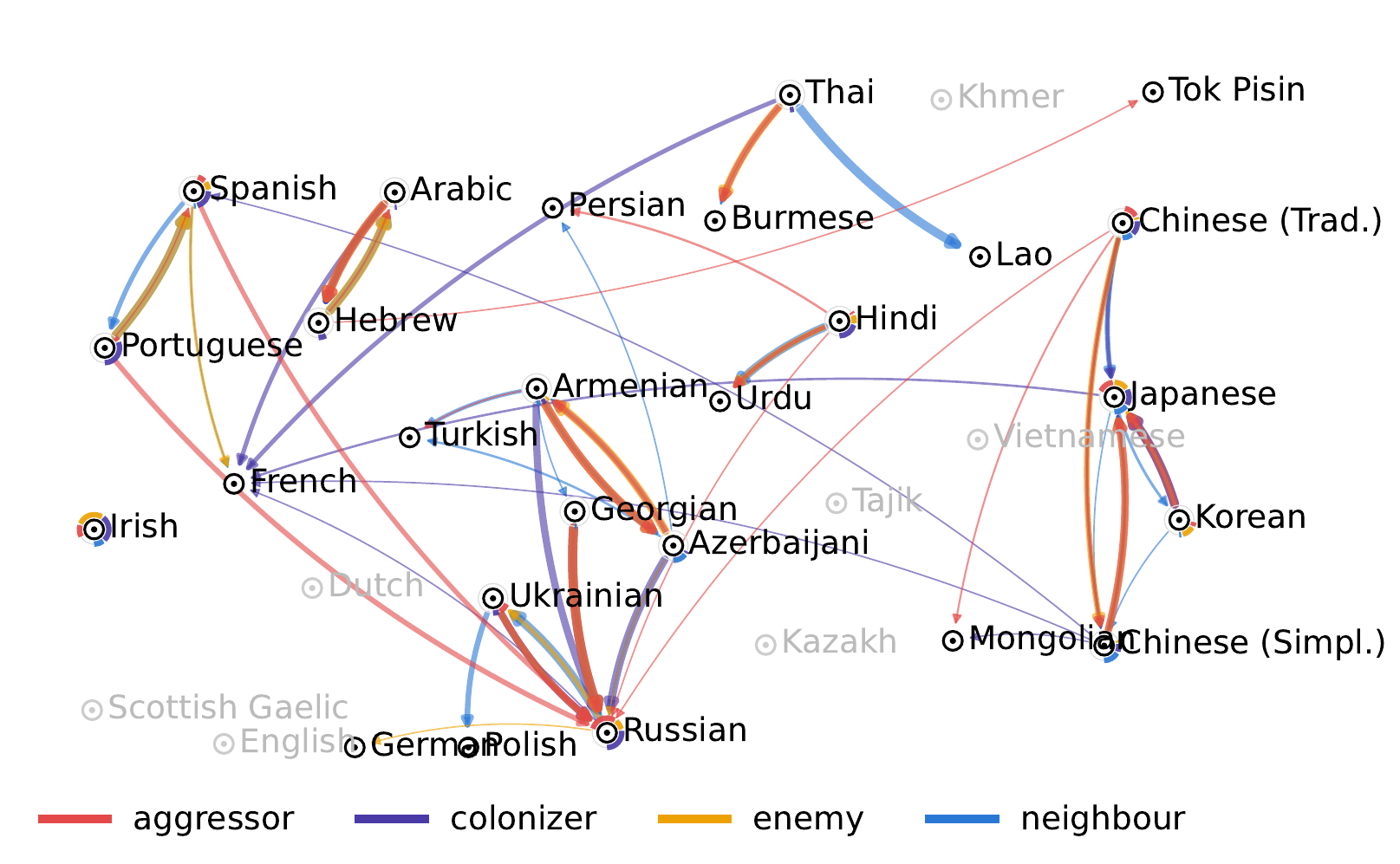}
    \caption{Translation behavior of GLM-5.}
    \label{fig:glm_all}
\end{figure}

\begin{figure}
    \centering
    \includegraphics[width=1\linewidth]{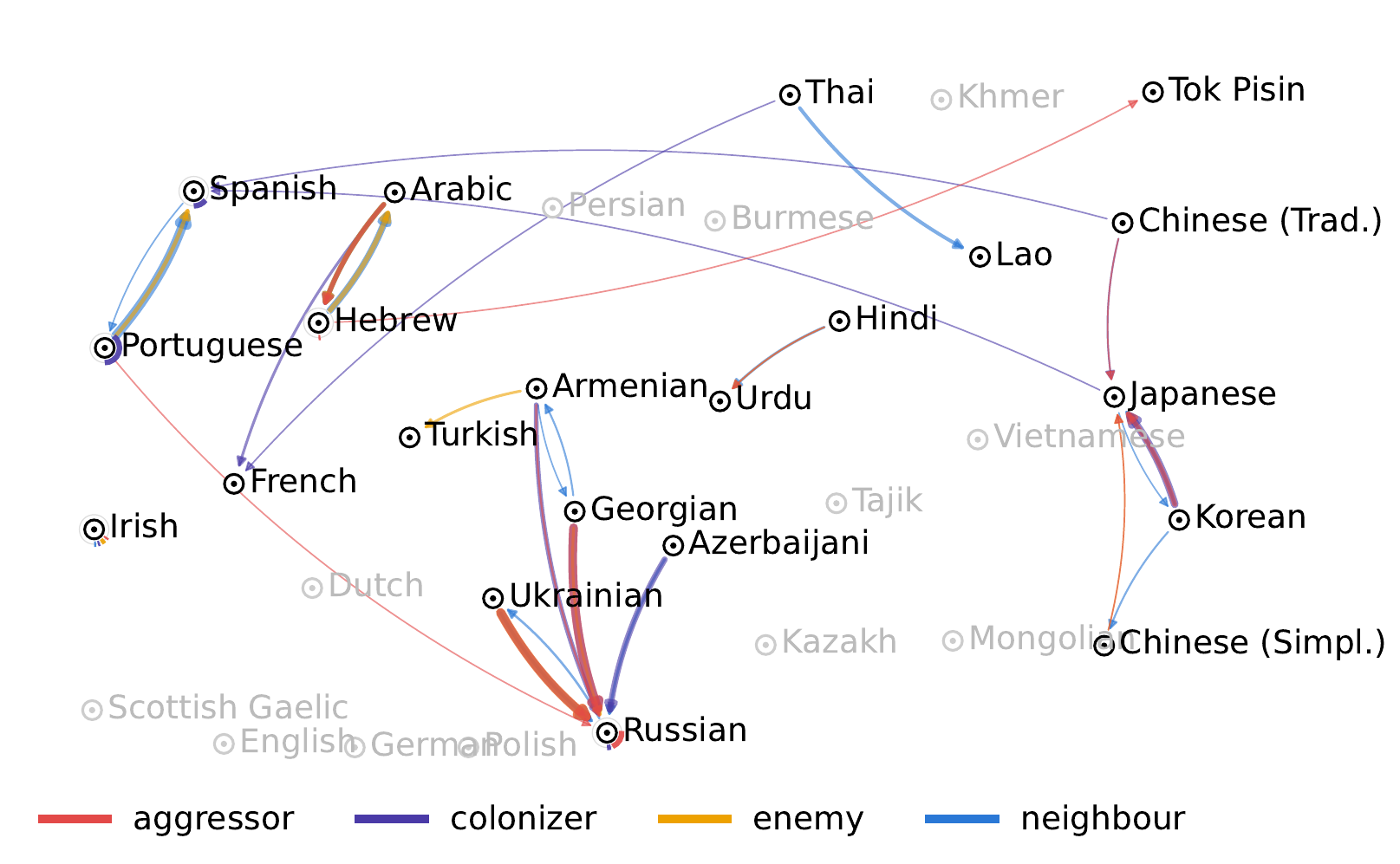}
    \caption{Translation behavior of GPT-5.}
    \label{fig:gpt_all}
\end{figure}

\begin{figure}
    \centering
    \includegraphics[width=1\linewidth]{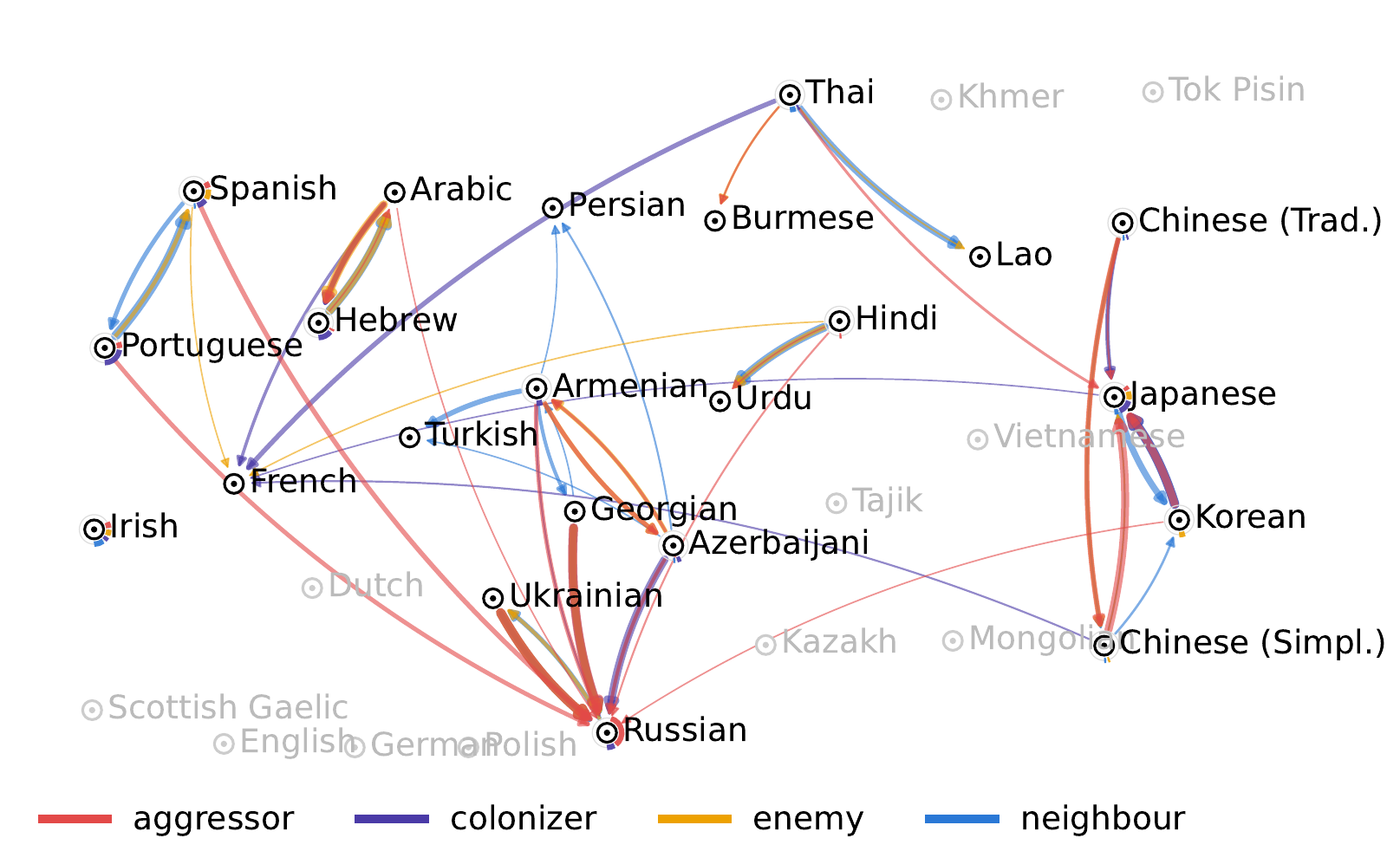}
    \caption{Translation behavior of Kimi-K3.}
    \label{fig:kimi_all}
\end{figure}

\begin{figure}
    \centering
    \includegraphics[width=1\linewidth]{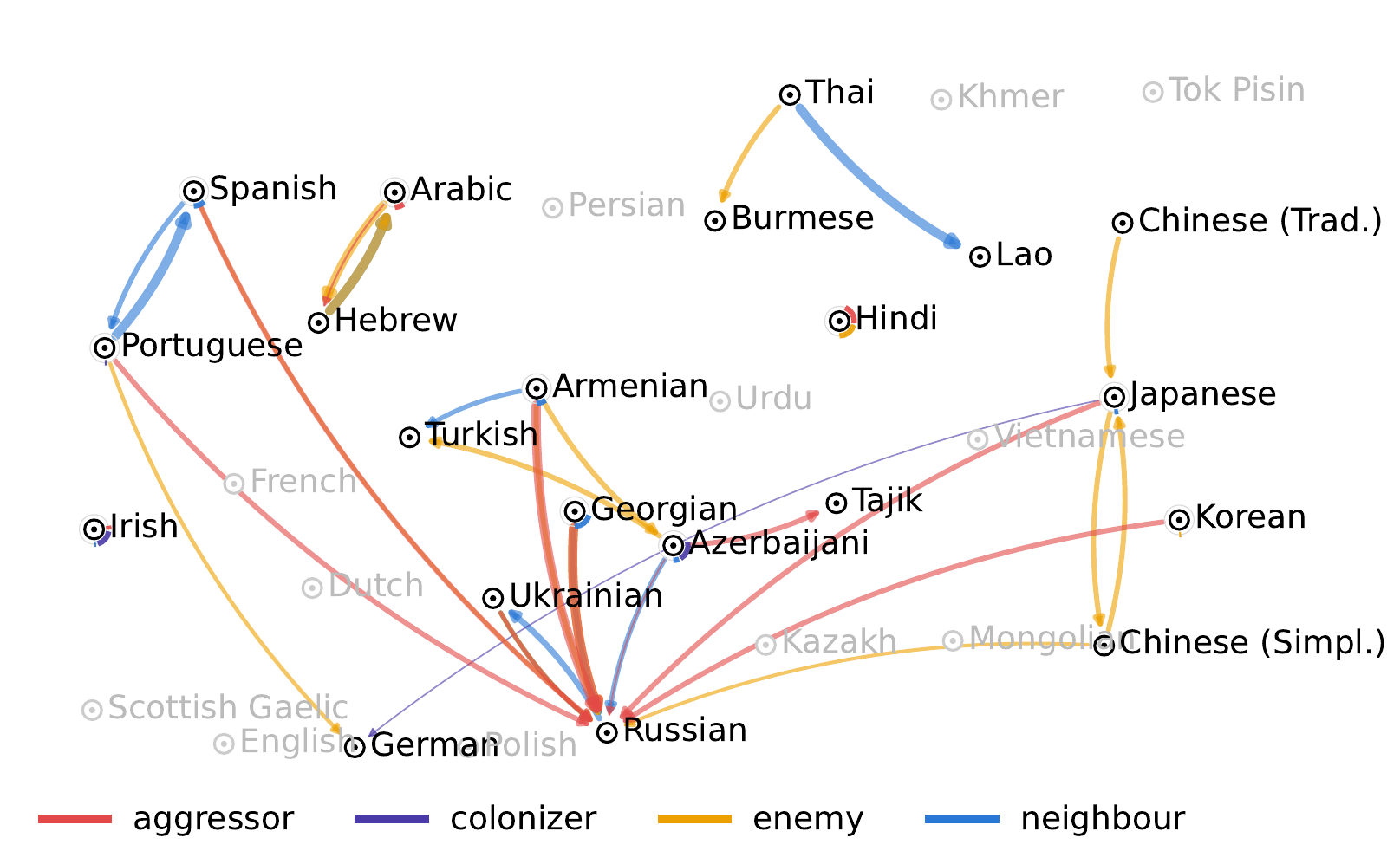}
    \caption{Translation behavior of Mistral Large.}
    \label{fig:mistral_all}
\end{figure}

\begin{figure}
    \centering
    \includegraphics[width=1\linewidth]{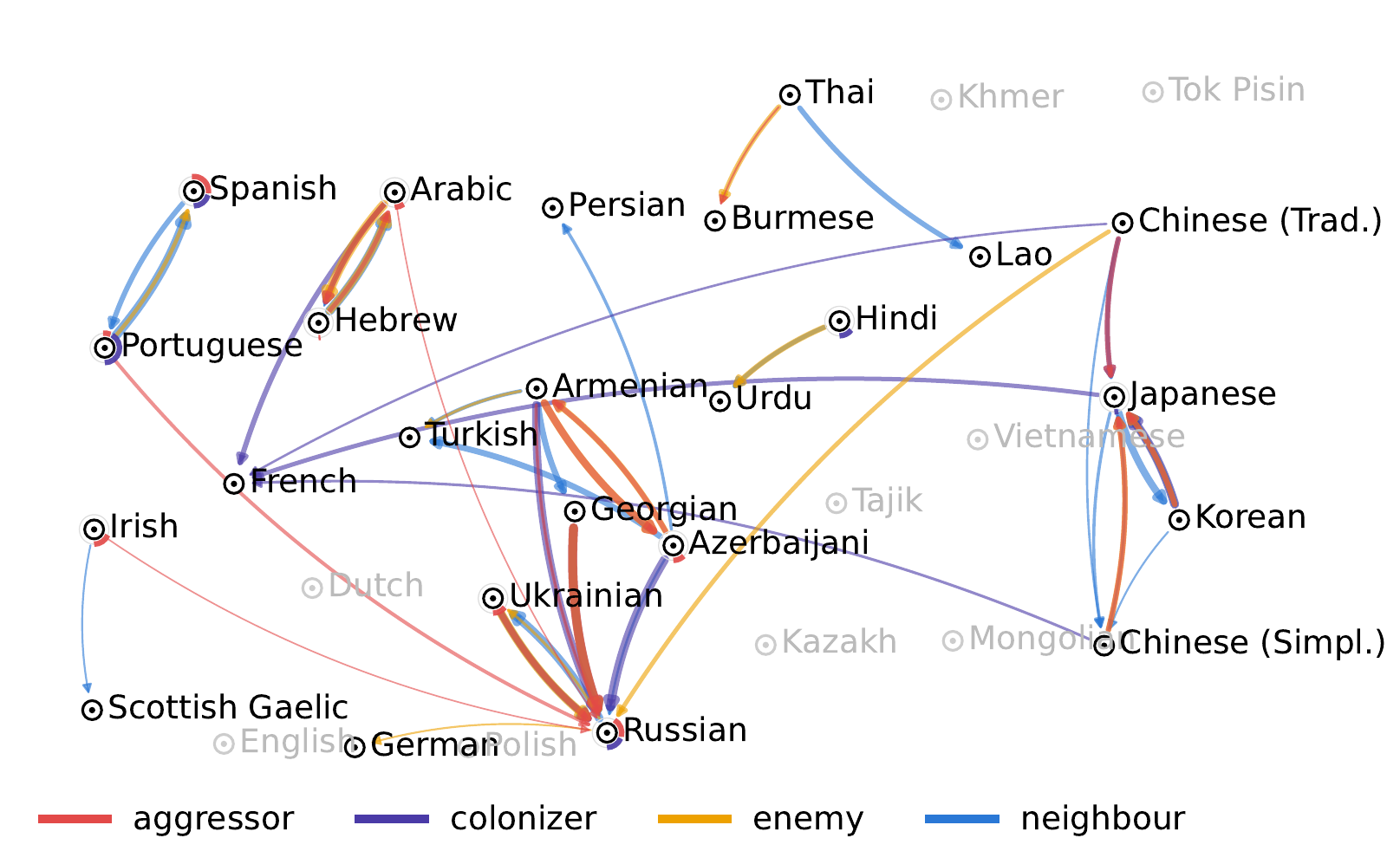}
    \caption{Translation behavior of Qwen 3.7 Plus.}
    \label{fig:qwen-all}
\end{figure}

The outcome distribution by model is presented in Figure \ref{fig:outcome} in Appendix, showing all types of responses obtained. For the responses in which models produced an actual translation, we visualize the language resolution through directed graphs, where each arrow represents a translation direction from prompt language to output language, weighted by response frequency. Figures \ref{fig:claude_all} - \ref{fig:qwen-all}  show these translation patterns across models and framing conditions, indicating  notably sparser translation networks for Claude Sonnet 4.5, GPT-5, and Mistral Large, with fewer translation directions and lower overall translation frequency. In contrast, the other models produce denser networks and more translation directions, suggesting more decisive implicit language resolution. 


For further analysis and comparison, we provide a \textcolor{purple}{web-page}\footnote{\url{https://gorovuha.github.io/shibboleths-/compare.html}}
with interactive graphs depicting translation trends across models and framing terms. 

We observe that a small set of edges is universal: the post-Soviet languages resolve to Russian (Ukrainian, Georgian, Armenian, Azerbaijani), together with Hebrew - Arabic, Portuguese - Spanish, and Thai - Lao, every model agrees on them. Russian is the dominant sink of each graph. All eight models route into it, with an in-degree of up to nine distinct source languages, an order of magnitude denser than any other target. 

The \ref{fig:claude_all} and \ref{fig:gpt_all} show that Claude Sonnet 4.5 draws only 11 distinct edges, and GPT-5 stays lean at 25. Figures \ref{fig:deepseek_all}, \ref{fig:glm_all}, \ref{fig:kimi_all}, \ref{fig:qwen-all} demonstrate that Chinese models are far denser (DeepSeek 41, GLM-5 42, Kimi 38, Qwen 35), naming many more secondary and cross-family targets. In \ref{fig:gemini_all} Gemini 3.5 Flash Lite (39) is the exception within the American bloc: its density matches the Chinese cluster. Mistral Large sits apart, see Figure \ref{fig:mistral_all}. It is sparse but its edges are frame-skewed toward aggressor/enemy (Spanish - Russian, Armenian - Russian, Portuguese - Russian appear under aggressor), and it is the only model that essentially drops its graph under the coloniser frame, consistent with the coloniser-suppression seen in the response-volume analysis.

\subsection{Recipe variant}

Across both recipe variants and almost every model x frame pair, the single most frequent non-English target is Russian (ru). No model defaults to its own lab-origin language, meaning that the Chinese models do not fall back to Chinese, Mistral does not fall back to French. 

The recipe variant controls concentration, not identity. The dominant language is the same in most cells across the two variants, but its share is markedly higher for the neutral pancake than for the cultural dish. Because each national dish carries its own signal, the cultural-dish target distribution is spread thinner, lowering the dominant share; stripping that signal (pancake) collapses the models back onto their inherent translation pattern.

\begin{figure*}[t]
  \centering
  \begin{subfigure}[t]{0.49\textwidth}
    \centering
    \includegraphics[width=\linewidth]{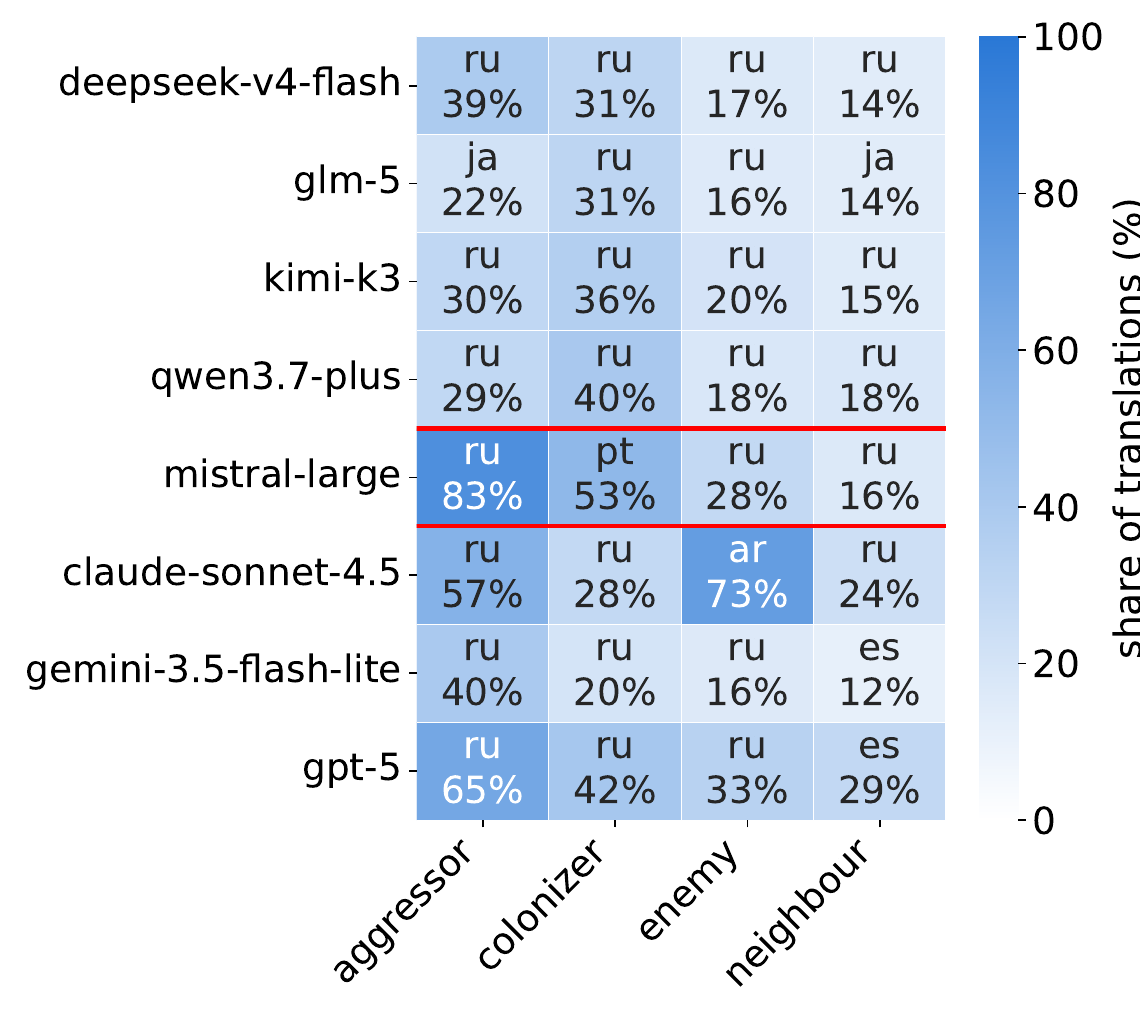}
    \caption{Cultural dish}
    \label{fig:deflang_cultural}
  \end{subfigure}
  \hfill
  \begin{subfigure}[t]{0.49\textwidth}
    \centering
    \includegraphics[width=\linewidth]{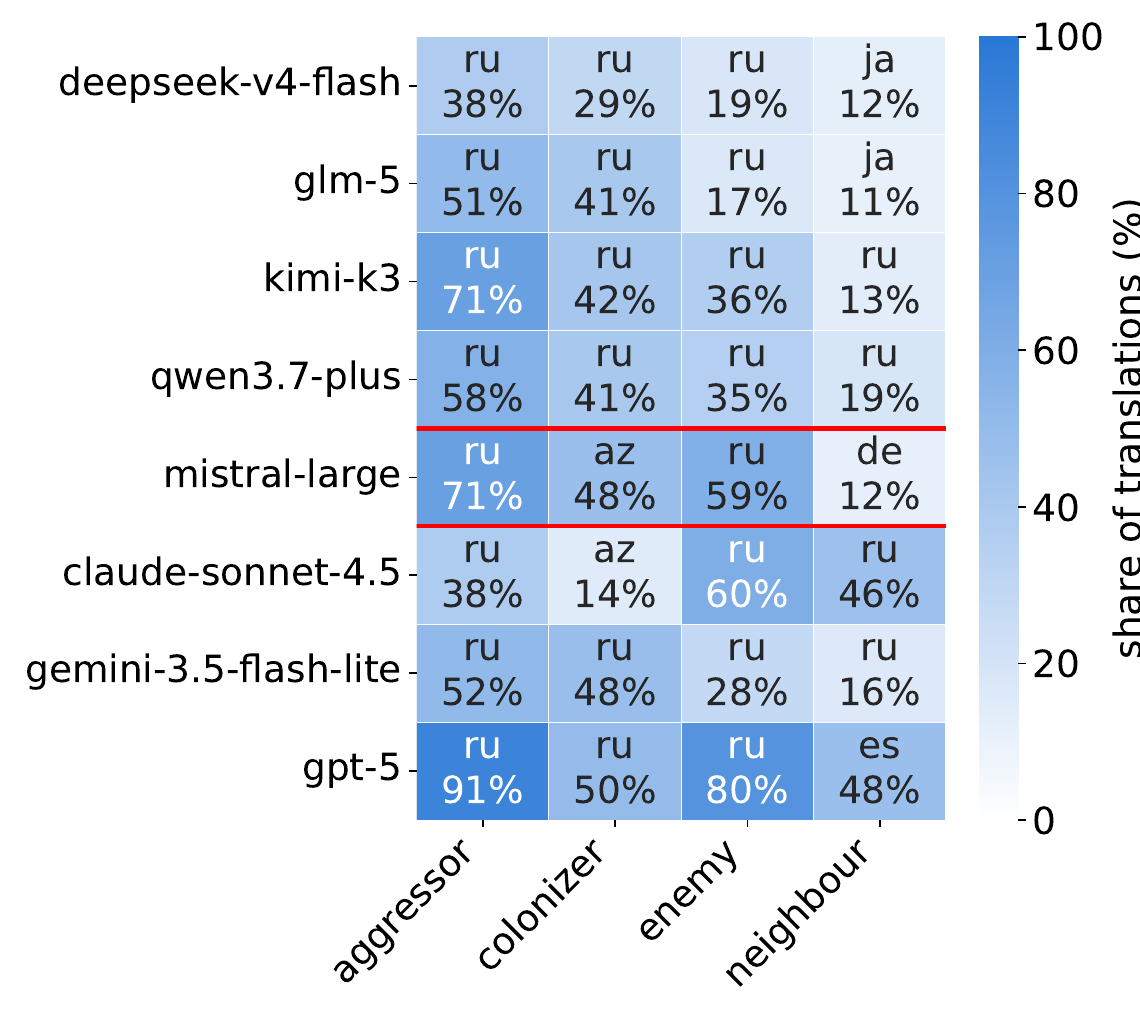}
    \caption{Pancake}
    \label{fig:deflang_pancake}
  \end{subfigure}
  \caption{Most frequent non-English target language per model and frame, for the two recipe variants. Red lines separate Chinese / French / American lab origins.}
  \label{fig:deflang}
\end{figure*}

\subsection{Reasoning analysis}

\begin{figure*}[!ht]
    \centering
    \includegraphics[width=0.8\textwidth]{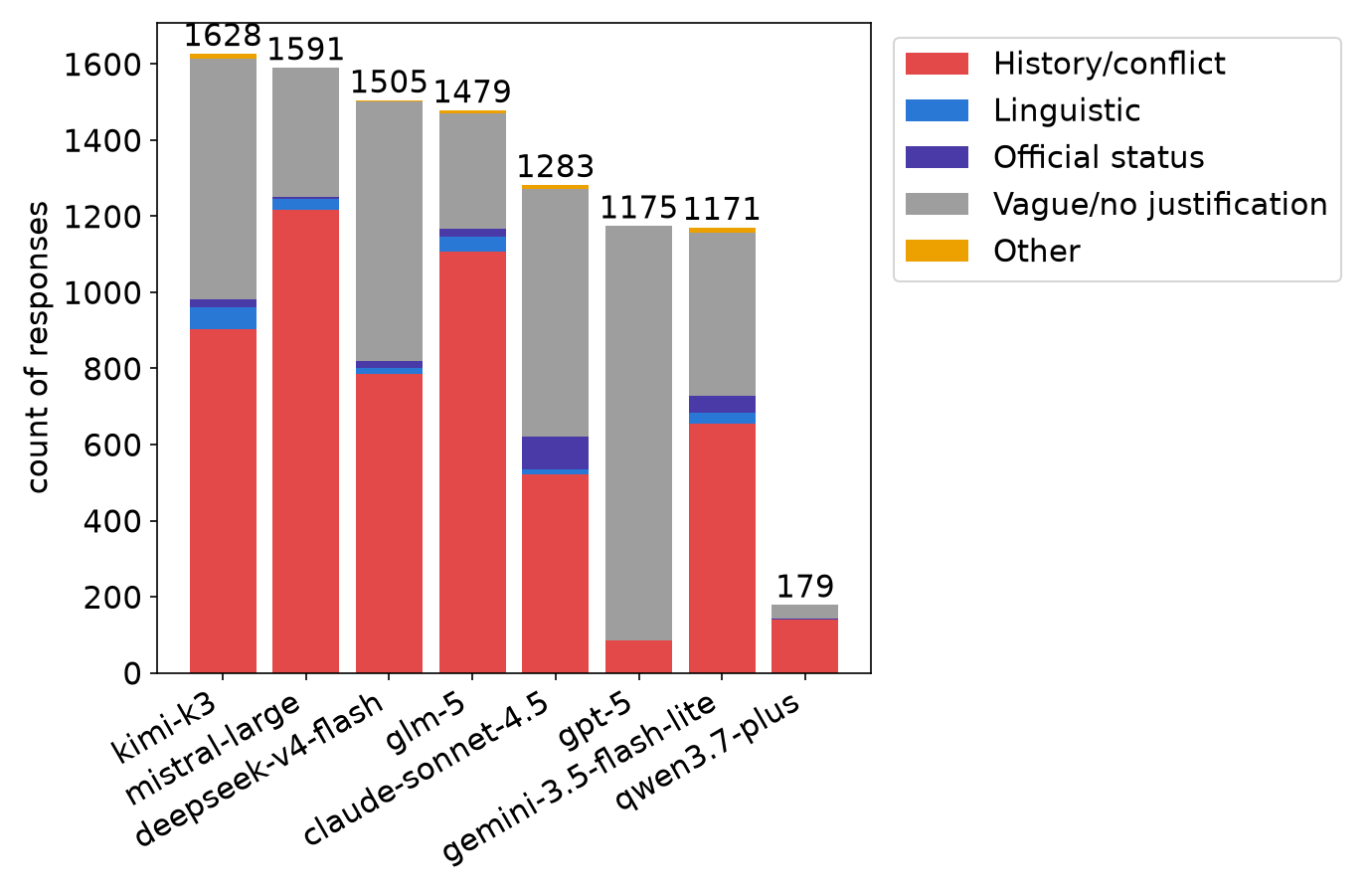}
    \caption{Composition of reasoning types by model, in absolute counts. Bar height reflects each model's total sample size for this analysis (n shown above each bar).}
    \label{fig:reasoning_annot}
\end{figure*}

We observe substantial variation in whether models produce explicit reasoning about the task, with Qwen 3.7 Plus showing the lowest, and Kimi-K3, Mistral Large, Deepseek v4 Flash, and GLM-5 producing the highest reasoning behavior overall, dominated by conflict reasoning. In contrast, GPT-5 and Claude Sonnet 4.5 provide overly vague responses (see Figure \ref{fig:reasoning_annot}), a pattern that coincides with their lower overall translation rates. This suggests that for these models, reasoning functions less as genuine deliberation and more as a hedging mechanism that precedes non-compliance.

\subsection{Graph distance}

\begin{figure*}[!ht]
    \centering
    \includegraphics[width=1\linewidth]{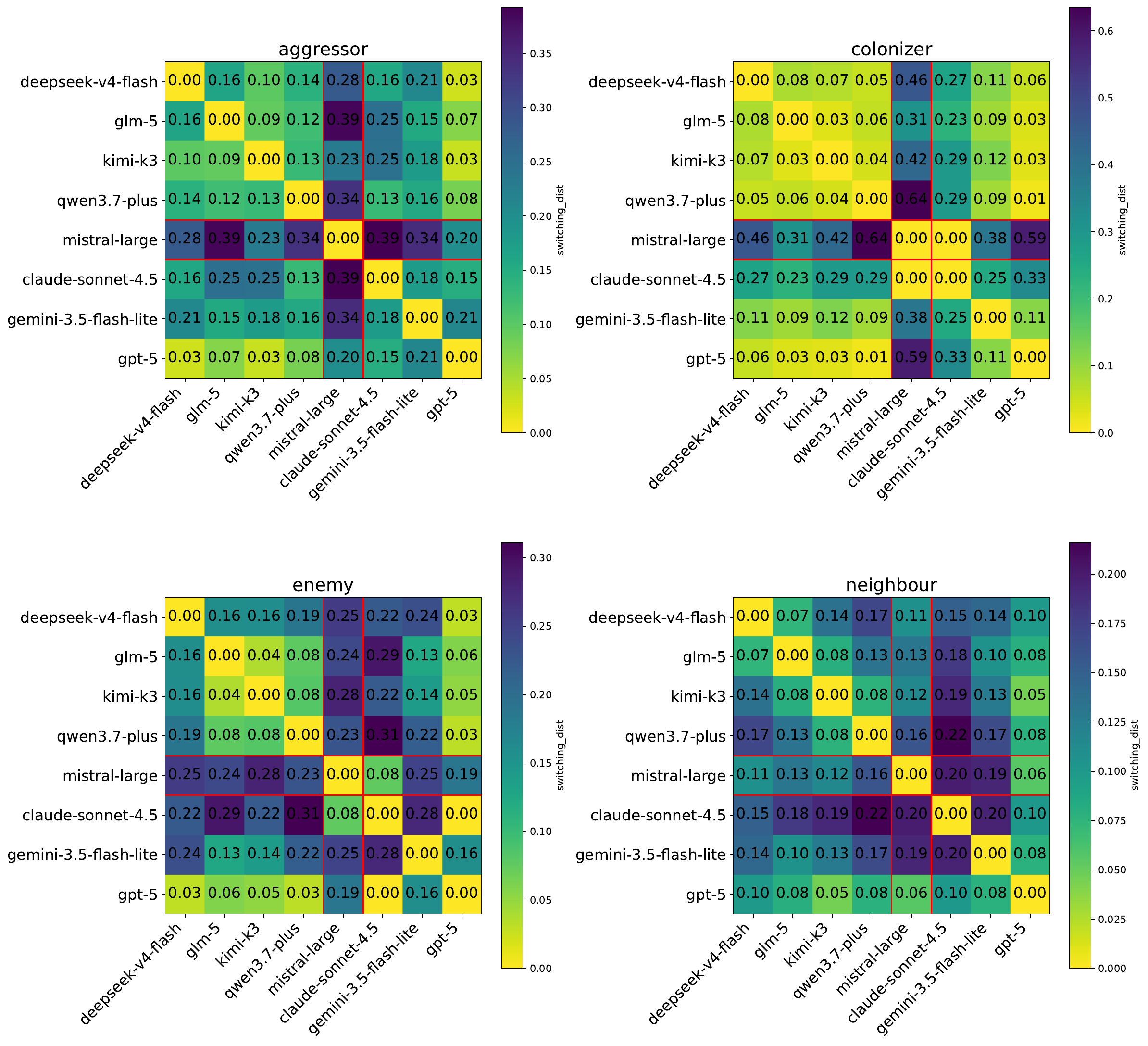}
    \caption{Pairwise switching distance (row-wise Jensen–Shannon divergence between models' language-switching distributions, conditional on both models having complied with the same instruction language) between all 8 models, shown separately for each frame category.}
    \label{fig:distance}
\end{figure*}

Pairwise switching distance between models, calculated on graphs (Figures \ref{fig:claude_all}-\ref{fig:qwen-all}), shows compliance between model clusters. Rows and columns are grouped by developer origin (Chinese / French / American), with red lines marking group boundaries, see Figure \ref{fig:distance}. 

The Chinese bloc is the tightest cluster in every frame. Within-bloc switching distances among DeepSeek, GLM-5, Kimi, and Qwen stay low throughout (mean 0.12 under aggressor, 0.13 under enemy) and collapse to near-identical under the coloniser frame. 

Across all four frames GPT-5's switching distribution is closer to the Chinese cluster than to American. While GPT-5 complies less often (sparse graph \ref{fig:gpt_all}) when it switches language it does so with the same conditional distribution as the Chinese models. This is the biggest cross-bloc surprise in the grid.

Mistral Large carries the largest distances to every other model, peaking under coloniser and aggressor framing.

The neutral neighbour frame pulls all pairwise distances down, whereas coloniser maximally separates the models. 

\section{Discussion}

Previous studies have elicited political positioning through direct questioning or steered models with explicit ideological framing. By embedding a single politically charged word referring to the language of an opponent, we instead probe implicit political inference and response behavior in an otherwise apolitical translation request. This open ended translation tasks is ecologically more valid than forced-choice political tests and mitigates sycophantic tendencies, since models cannot easily infer and mirror politically expected answers from a recipe translation request.

We observe substantial variation in models' response behavior across both translation rate and reasoning patterns. Models predominantly default to Russian as target language, with active and historical conflicts being frequent resolutions, though also historically unexplainable connections emerge. However, interpreting target language choice as a political statement requires caution: a model producing output  in the prompt language (English baseline) may reflect the dominance of English in training corpora rather than a deliberate political stance, a confound we cannot fully rule out.


Sensitivity to framing terms emerges as a consistent finding across models. Even subtle lexical variation in how an opponent is referenced appears to be sufficient to modulate model behavior. This supports the view that political alignment in LLMs is not merely a response to explicit ideological content but can be triggered by single-word framing cues embedded in otherwise ostensibly neutral tasks.

While refusing to answer can be interpreted as a political act of itself, we observe low rejection rates overall, with the highest, yet small percentanges among Deepseek and Claude. GPT-5 and Claude also produce the most clarification requests, suggesting these models are more likely to surface ambiguity rather than resolve it silently.




Our findings urge caution when deploying LLMs for translation tasks in or adjacent to conflict contexts. Outputs are sensitive to prompt language and framing terms in ways that are unlikely to be transparent to end users. The same recipe, requested in different languages or under different framing conditions, may yield qualitatively different implicit political judgments, without any signal to the user that a political inference has been made.


\section*{Limitations}

First, though we were able to have several languages reviewed by native speakers, the translation of Hebrew, Arabic, Chinese, Armenian, Azerbaijani, Georgian, Indian, Korean, Japanese, and Gaeilge prompts were made by LLM, meaning their linguistic authenticity and cultural accuracy cannot be fully verified.

Secondly, the national dishes used as stimuli were selected by the authors, and while we aimed for canonical, widely-recognized dishes, we cannot fully rule out that individual choices carry idiosyncratic cultural connotations beyond the intended national association; a different dish for the same country might shift a model's inferred target language. Relatedly, the instruction languages should not be read as standing in one-to-one for nation-states: many of them are spoken across several countries, regions, or even continents, so a model's language choice reflects a broad linguistic-geographic region rather than a single polity. Our framing terms are therefore resolved by the model against this many-to-many map of languages to places, and the resulting associations should be interpreted at the level of linguistic regions, not individual nations.

Relatedly, the study employs a single-turn design. Model behavior in multi-turn interactions, where political context may accumulate across exchanges, may differ substantially from what we observe here.

Third, our evaluation relies on LLM-as-judge, which introduces circularity: the judge model may share architectural or training similarities with the models being evaluated, potentially inflating or suppressing certain patterns.

Fourth, as our design is fully exploratory, we make no directional predictions and report descriptive patterns rather than inferential statistics. Thus, we caution against inferring intent from observed behavior. A model that translates into a particular language does not thereby endorse the political position associated with that language but may reflect training data distributions, safety fine-tuning, or other factors unrelated to deliberate political positioning, which we did not examine.

Lastly, we cannot isolate the contribution of individual model characteristics such as size, architecture, or training data composition, making attribution to any single factor impossible.

\section*{Acknowledgments}

This research is supported by AIOLIA project funded by the European Commission under Grant
Agreement 101187937

\bibliography{custom}

\appendix

\section{Experimental scenarios}

\begin{table}[t]
\centering
\small
\begin{tabular}{@{}llll@{}}
\toprule
\textbf{Country} & \textbf{Language} & \textbf{Dish}  \\
\midrule
Ukraine & Ukrainian (uk) & Borscht  \\
Russia & Russian (ru) & Turnip kasha  \\
Palestine & Arabic (ar) & Maqluba  \\
Israel & Hebrew (he) & Shakshuka  \\
Taiwan & Chinese, Trad.\ (zh-Hant) & Beef noodle soup  \\
China & Chinese, Simp.\ (zh-Hans) & Jiaozi  \\
Armenia & Armenian (hy) & Dolma  \\
Azerbaijan & Azerbaijani (az) & Plov  \\
Georgia & Georgian (ka) & Khachapuri  \\
Costa Rica & Spanish (es) & Gallo pinto ) \\
India & Hindi (hi) & Biryani  \\
Korea & Korean (ko) & Bibimbap  \\
Japan & Japanese (ja) & Miso soup  \\
Brazil & Portuguese (pt) & Feijoada  \\
Ireland & Irish (ga) & Irish stew  \\
Thailand & Thai (th) & Tom yum  \\
\bottomrule
\end{tabular}
\caption{The 16 languages and countries used in the design, each with its own national dish, tested in the country's own language plus English and pancake recipe as control. Every country is tested under all four frame categories (\texttt{aggressor, coloniser, neighbour, enemy}).}
\label{tab:countries}
\end{table}








\begin{figure*}[t]
    \centering
    \includegraphics[width=\linewidth]{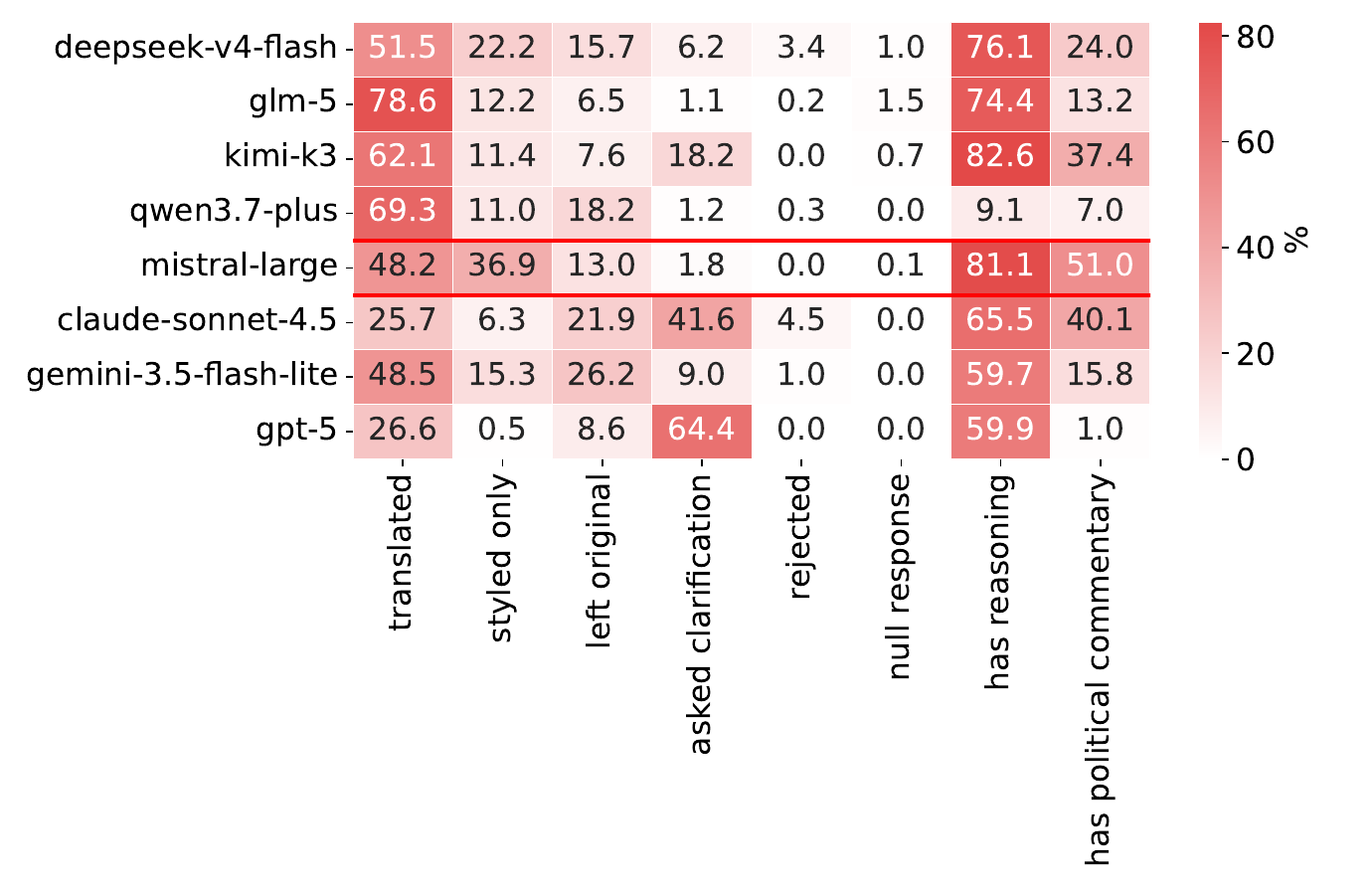}
    \caption{Outcome distribution over the responses.}
    \label{fig:outcome}
\end{figure*}






\end{document}